\def\year{2021}\relax
\documentclass[letterpaper]{article} 
\usepackage{aaai21}  
\usepackage{times}  
\usepackage{helvet} 
\usepackage{courier}  
\usepackage[hyphens]{url}  
\usepackage{graphicx} 
\usepackage{natbib}  
\usepackage{caption} 
\usepackage{xcolor}
\usepackage{multirow}
\usepackage{makecell}
\usepackage{nicematrix}

\newcommand{\sysname}{{\sc NovGrid}}

\title{NovGrid: A Flexible Grid World for Evaluating Agent Response to Novelty}

\author {
Jonathan Balloch,\textsuperscript{\rm 1*} 
Zhiyu Lin,\textsuperscript{\rm 1}
Mustafa Hussain,\textsuperscript{\rm 1} 
Aarun Srinivas,\textsuperscript{\rm 1} 
Robert Wright,\textsuperscript{\rm 2} 
Xiangyu Peng,\textsuperscript{\rm 1} 
Julia Kim,\textsuperscript{\rm 1} 
Mark Riedl\textsuperscript{\rm 1}\\
}
\affiliations {
    \textsuperscript{\rm *} balloch@gatech.edu 

    \textsuperscript{\rm 1} 
    Georgia Institute of Technology \\
    Atlanta, GA, USA \\

    \textsuperscript{\rm 2} 
    Georgia Tech Research Institute \\
    Atlanta, GA, USA \\

}

\begin{document}

\maketitle

\begin{abstract}
A robust body of reinforcement learning techniques have been developed to solve complex sequential decision making problems. 
However, these methods assume that train and evaluation tasks come from similarly or identically distributed environments.
This assumption does not hold in real life where small novel changes to the environment can make a previously learned policy fail or introduce simpler solutions that might never be found.
To that end we explore the concept of {\em novelty}, defined in this work as the sudden change to the mechanics or properties of environment.
We provide an ontology of for novelties most relevant to sequential decision making, which distinguishes between novelties that affect objects versus actions, unary properties versus non-unary relations, and the distribution of solutions to a task.
We introduce \sysname, a novelty generation framework built on MiniGrid, acting as a toolkit for rapidly developing and evaluating novelty-adaptation-enabled reinforcement learning techniques.
Along with the core \sysname we provide exemplar novelties aligned with our ontology and instantiate them as novelty templates that can be applied to many MiniGrid-compliant environments.
Finally, we present a set of metrics built into our framework for the evaluation of novelty-adaptation-enabled machine-learning techniques, and show characteristics of a baseline RL model using these metrics.
\end{abstract}

\section{Introduction}

There exists a robust body of machine learning techniques--including but not limited to imitation learning and reinforcement learning--that can be used to form learning models of agent behavior in complex sequential decision making environments. 
These techniques can be generally applied to find an optimal policy that solves nearly any problem that can be modeled as a Markov Decision Process, and the policies can be anything from simple look-up tables to Gaussian Processes and Deep Neural Networks~\cite{engel2005reinforcement,sutton2018reinforcement}. 

However, success in these learning methods shares a common assumption: the stochastic process used to model the environment is equivalent in both training and evaluation.
While this train-test similarity assumption holds in some settings, in most real-world cases environments cannot ever be guaranteed to function the same forever. 
Whether in environments related to health care---the industry which the United States Bureau of Labor Statistics projects to have the greatest projected labor demand over the next decade---or freight driving which is an integral part of the modern supply chain, agents encounter and need to respond to \textit{novelty} \cite{usbls_2021}. 
To eventually meet these real-world challenges, the reinforcement learning research community needs to analyze how well different agents response to a wide variety of novelties. 

\begin{figure}[t]
    \centering
    \includegraphics[width=\linewidth]{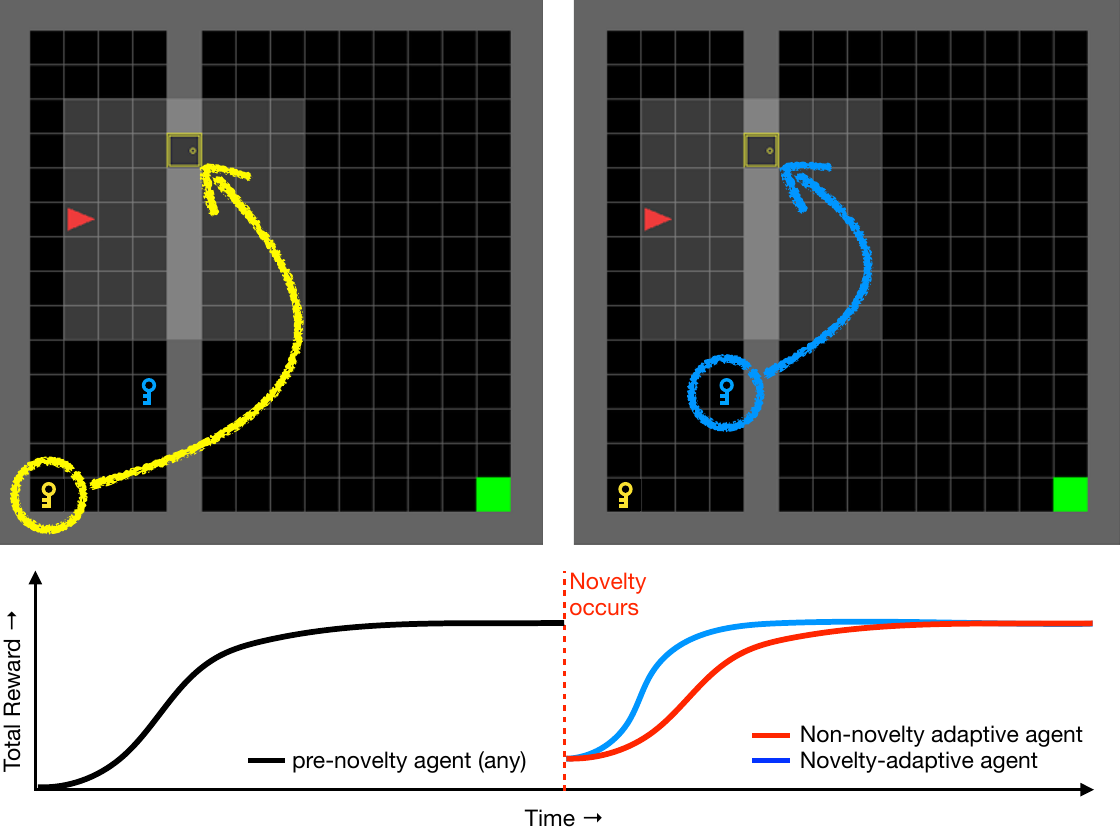}
    \caption{Illustrative example of \sysname{}. Initially the yellow key opens the door so the agent (red triangle) can get to the goal (green box). 
    The agent learns a converged policy.
    At a certain time, the yellow key stops working and the blue key opens the door.
    The agent's performance drops off and recovers (bottom).
    The blue and red lines are notional learning curves for agent with and without novelty adaptation, respectively. }
    \label{fig:splash}
\end{figure}

We provide three contributions.
First, we propose an ontology of novelty for sequential decision making that distinguishes between object novelties (new or changed properties of objects) and action novelties (changes in how the agent's actions work).
Our ontology also relates novelties to goal-seeking performance, categorizing novelties as to whether they hinder or facilitate future expected reward.
Second, we introduce {\sc Novelty MiniGrid} (\sysname), an extension of MiniGrid environment~\cite{gym_MiniGrid} that allows for the world properties and dynamics to change according to a generalized novelty generator.
The MiniGrid environment is a grid-world that facilitates reinforcement learning algorithm development with low environment integration overhead, which allows for rapid iteration and testing.
\sysname{} extends the MiniGrid  environment by expanding the way the grid world and the agent interact to allow novelties to be injected into the environment.
Specifically this is done by expanding the functionality of the actionable objects---doors, keys, lava, etc.---already in MiniGrid and creating a general environment wrapper that injects novelty at a certain point in the training process.
We provide a number of example novelties aligned with different dimensions of our novelty ontology (in addition to allowing developers to create their own novelties).
Third, we propose a set of metrics important to measuring adaptation to novelty for the evaluation of agents. 
With these components \sysname{} will enable more rapid research progress on agent novelty adaptation.  

In this paper we will first provide a background on novelty adaptation in sequential decision making problems and its relationship to prior work. 
We will then discuss the ontology of novelties organized by the different characteristics we understand as most important to sequential decision making agents. 
After this, we detail the ways \sysname{} environment was designed and how the provided novelty implementations enable research on each part of this ontology. Finally we describe the metrics used to test novelty and some sample performance of the system. 

\section{Novelty Background and Related Work}

In this section we discuss some of the background and motivation for novelty adaptation as a research challenge.

Novelty can be described in the context of differences in an environment over a period of time and associated capabilities to detect and respond to those changes~\cite{langley2020open}. 
Boult et al.~\shortcite{boult2021towards} looks to unify the study of novelty both sequential decision making and traditional machine learning domains, categorizing novelties broadly as \textit{world novelties}, \textit{agent novelties}, and \textit{observation novelties}. 
By Boult et al. definition, world novelties are changes in the objects and dynamics of the world external to the agent as well as the agent's abilities to affect change on the world.
Agent novelties are those where the agent's state does not align with the agent's prior understanding of the world and/or can be classified by the agent as a novelty. 
Observation novelties are those where a sense-making tool external to the agent (e.g., a sensor) is subject to changes in the environment, such as when a camera or radar signal experiences deterioration from use or unexpected interference. 

It is important to note concepts and research areas that should {\em not} be confused with or included in novelty.
Novelty and novelty adaptation is not equivalent to methods related to outliers like outlier detection or rejection---outliers assumes \textit{a priori} some correct or expected distribution, where the points or behaviors that lie outside that distribution are aberrant behavior.
By contrast, novelty is a sudden change that changes the environment distribution unexpectedly~\shortcite{chandola2007outlier}.

Novelty adaptation is also different from {\em continual learning} or {\em lifelong learning}.
In continual and lifelong learning there exists a sequence "tasks", each of which could be different environments, datasets, or novel classes. 
These tasks can be overlapping, task boundaries don't have to be well-defined, and can include a mixture of supervised and unsupervised data, but in most cases these tasks are disjoint and task boundaries are known and discrete~\cite{parisi2019continual,silver2013lifelong,smith2021memory}. 
Most distinctly, in continual learning the model is trained on only one task at a time but validated on that task and all prior tasks. 
Novelty adaptation, on the other hand, only requires the agent to perform well at the task at hand, so before the introduction of the novelty the agent is evaluated only on the pre-novelty world and then only on post-novelty world after novelty is introduced.

Novelty adaptation is a superset of domain shift and domain adaptation. 
Domain adaptation in sequential decision making problems specifically addresses problems where training and deployment have the same feature space but different distributions over that feature space~\cite{zhang2019vr,sun2015survey}. 
Additionally, this research domain usually assumes that the agent must adapt to this extremely quickly--few shot domain adaptation--or simply be robust to these changes--as in zero-shot domain adaptations. 
By this definition, domain shift and domain adaptation are one aspect of novelty and novelty adaptation, but novelty goes beyond this by also including variation of the fundamental dynamics of an environment and the agent interactions as well. 


The closest analog to novelty adaptation is transfer learning. Given a set of source or training tasks and a target task, transfer learning aims to learn an optimal policy for the target domain leveraging what it learned from the source domains as well as what it has access to in the target domain \cite{zhu2021transfer}. 
While there are some works that focus on variants like zero-shot transfer learning--most notably research focused on transferring from a simulation to the real world--for the most part transfer learning is studies the mechanics of model reuse and fine tuning~\cite{higgins2017darla}. 
That is to say: given an already trained model trained on a certain task, how would one reuse this model in a new task where this new task can range from being a subtask of the original task to being completely unrelated. 
In novelty adaptation on the other hand is the pre-novelty and post-novelty tasks are always related by a (usually realistic) transformation. 
While transfer learning optimizes the reuse of a learned model with no constraints on the relationship between source and target tasks and by any means necessary, novelty adaptation optimizes for adaptation performance and efficiency online given the knowledge that a realistic transformation between source and target exists. 

We are not the first to investigate the characterization and evaluation of novelty adaptation in sequential decision making problems. Pimentel et. al.~\shortcite{pimentel2014review} conducted a comprehensive survey of novelty detection that characterized novelty types.
There has also been a small amount of research that studies how agents might more effectively adapt. Approaches range from adaptive mixed continuous-discrete planning to knowledge graphs have been used in combination with actor-critic reinforcement learning techniques to improve both detection and adaptation \cite{klenk2020model,peng2021detecting}.
Recently there have even been strong efforts to formulate a unified theory of novelty detection and novelty characterization, and to conceive a metric with which the degree of all novelties can be measured \cite{boult2021towards,langley2020open,alspector2021representation}. There also has been people who have worked on problems extremely relevant to the space of novelty and novelty adaptation without being aware of the novelty adaptation space, such as work on adaptation ``hidden'' domain shifts \cite{chen2021active}.

Most similarly to this work there has been recent work on environments for novelty detection, characterization, and adaptation in sequential decision making problems.
The Novgridworld environment implements a Minecraft-inspired grid world to study novelties in the context of agent-driven object-object interaction \cite{goelnovelgridworlds}.
The GNOME Monopoly environment examines multi-agent gameplay in a long term multi-faceted strategic context, while Science Birds has a fewer number of timesteps and in each episode by examining both observational and world novelties \cite{kejriwal2021multi,gamagenovelty}. In both cases, an implementation is provided for outside users. 
These works set a firm foundation for studying novelty, but what \sysname{} uniquely provides are highly extendable implementations what can be applied to any MiniGrid environment with an OpenAI Gym interface, and it provides a standard set of novelties provided with the implementations making it easy for researchers to benchmark progress through time.

\section{An Ontology of Novelty for Sequential Decision Making}\label{sec:ontology}

We model sequential decision making problems as having two fundamentally different types of entities: agents and the environment, which we model as as interacting in a stochastic game---a multi-agent generalization of Markov Decision Process (MDP). 
The injection of novelty constitutes a transformation from the original game or MDP $M$ to a new game or MDP $M'$. 
Given this model of environments, we consider all aspects of the problem except a agent's decision making model to be property of the environment. This includes agent morphology, sensors, and action preconditions and effects. 
As a result, the ontology we lay out here can be considered a specification of Boult et. al.'s world novelties in the context of sequential decision making problems.

With this model of the environment we assume that each agent's observation space and action space remains consistent before and after novelty is injected. 
That is, the number of actions and the size and shape of observations are consistent throughout each experiment.
That said, the manifestation of these fixed sets may change; actions that initially have some specific effect or no effect pre-novelty can take on different effects post-novelty.
Likewise, there may be observations and states that never occur pre-novelty that start to occur post-novelty.
This is consistent with a robotics perspective on MDPs where actions and observations are governed by an underlying physics of the real world, even though we experiment within grid worlds and games~\cite{kejriwal2021multi,gamagenovelty}.

We assume that the agent's mission $T$ is consistent before and after the novelty, meaning that we do not consider changes to the extrinsic agent rewards. 
While changes to the goal and reward structure of an agent are indeed important in the real world, this is closely related to continual lifelong learning and multitask learning. 
The integration of novelty adaptation with these fields is left to future work.

We characterize novelties along three dimensions. 
The first dimension is \textbf{object} vs \textbf{action} novelties.
Objects are any component of the environment that is not controllable. 
This includes keys, doors, balls, etc.
Object novelties involve changes to, or the introduction or removal of, objects or classes of objects. 
Actions are the ways in which the world is affected by controllable entities.
Action novelties involve changes in the dynamics of actions through which the state of the world is affected. 
Action novelties can involve changes in the preconditions of actions---the applicability criteria of actions---or action effects---the way in which the world is changed when an action is executed. 

Second, novelties can be expressed as changes to \textbf{unary} predicates or \textbf{non-unary} (or $n$-ary where $n>1$) relations.
Unary object novelties can be thought of as added, removed, or changes to intrinsic properties of objects like mass, volume, or shape. 
Non-unary object novelties are changes in the relationship between objects, which is to say properties of objects that are necessarily defined in the context of other entities.
Unary and non-unary action novelties involve 
(a)~the addition, removal, or change of properties of objects required for action applicability, or 
(b)~changes to the properties of objects or changes to the relationship between objects.

Third, we observe that novelties can be categorized according to how they change the distribution of solutions to a task: 
 \begin{itemize}
     \item {\bf Barrier novelty}---the optima in the solution distribution are longer after novelty than before novelty.
     For example: a door the agent must pass through to achieve a goal initially required one key pre-novelty but requires the agent to possess two keys simultaneously post-novelty.
     \item {\bf Shortcut novelty}---the optima in the solution distribution are on average shorter after novelty than before novelty.
     For example, a door that required a key pre-novelty then does not require any keys post-novelty.
     \item {\bf Delta novelty}---the optima in the solution distribution are the same before and after novelty injection.
     For example, a door that required one key pre-novelty then requires a different key post-novelty.
 \end{itemize}

\begin{table*}
\footnotesize
\centering
\begin{tabular}{c|c|c|c|c|}
\multicolumn{2}{c|}{} & {\bf Barrier} & {\bf Delta} & {\bf Shortcut}\\
\hline
\multirow{4}{*}{\bf Objects} & \multirow{2}{*}{\bf Unary} & {DoorLockToggle} & \multirow{2}{*}{GoalLocationChange} & DoorLockToggle\\
& & unlocked$\rightarrow$locked & & locked$\rightarrow$unlocked\\ \cline{2-5}
& \multirow{2}{*}{\bf Non-Unary} & {DoorNumKeys} & \multirow{2}{*}{DoorKeyChange} & \multirow{2}{*}{{ImperviousToLava}}\\
& & NumKeys=1$\rightarrow$NumKeys=2 & &\\
\hline
\multirow{4}{*}{\bf Actions} & \multirow{2}{*}{\bf Unary} & ActionRepetition & {ColorRestriction} & ActionRadius\\
& & PickCommands=1$\rightarrow$PickCommands=2 & YellowOnly$\rightarrow$BlueOnly & PickDistance=1$\rightarrow$PickDistance=2\\ \cline{2-5}
& \multirow{2}{*}{\bf Non-Unary} & TransitionDeterminism & \multirow{2}{*}{Burdening} & ForwardMoveSpeed\\
& & Deterministic$\rightarrow$Stochastic &  & ForwardStep=1$\rightarrow$ForwardStep=2\\
\hline
\end{tabular}
  \caption{Novelty Ontology Exemplars}
  \label{tbl:exemplars}

\end{table*}



\section{Novelty MiniGrid}

\sysname{} is built around an OpenAI Gym Wrapper and designed to be compatible with all MiniGrid environments. This means that \sysname{} additionally works as a jumping-off point to evaluate any of the many 3rd-party environments based on MiniGrid. This novelty adaptation package has three fundamental components: a novelty injection mechanism built  into the core wrapper class, new and modified objects and entities to work with the novelty ontology as we described, and the novelty generator as well as the sample novelties we have to exemplify our ontology. 

The core novelty injection system is designed to be simple so that it is applicable to as many MiniGrid environments as possible. The wrapper wraps the environment, and no arguments are required besides the environment, but users can also specify the \texttt{novelty\_injection\_episode}, the episode in which novelty is injected. 
Given a model in train mode, MiniGrid resets its grid at the beginning of every episode with the \texttt{reset} function being called against the environment, which in turn calls the function \texttt{\_gen\_grid}. 
Our novelty injection wrapper monitors the training cycle, and when the \texttt{novelty\_injection\_episode} is reached the wrapper class switches to using alternatives for the \texttt{reset} and \texttt{\_gen\_grid} functions.
Specifically, after the novelty injection episode, the system now uses \texttt{\_post\_novelty\_reset} and  \texttt{\_post\_novelty\_gen\_grid}. 
This allows the wrapper to quickly and easily load in and overwrite the old environment with the new one. 

As each novelty is different the \texttt{\_post\_novelty\_gen\_grid} method in the base NoveltyWrapper class is only an abstract method that acts as a template. 
Each implementation of a novelty for testing adaptation requires an implementation of a class that inherits from this NoveltyWrapper and implements the \texttt{\_post\_novelty\_gen\_grid} method. 
To exemplify both this process and the novelty ontology that we describe in Section \ref{sec:ontology} we have built 11 exemplar novelties that together cover all of the different categories of our ontology. 
This way all researchers using \sysname{} can test their agent's adaptation sensitivity to different parts of the novelty ontology. The novelties delivered with \sysname{} and how the respective objects would usually work in MiniGrid are:

\begin{itemize}
    \item \textbf{GoalLocationChange}: This novelty changes the location of the goal object. In MiniGrid the Goal object is usually at fixed location.
    \item \textbf{DoorLockToggle}: This novelty makes a door that is assumed to always be locked instead always unlocked and vice versa. In MiniGrid this is usually a static property.
    If a door that was unlocked before novelty injection is locked and requires a certain key after novelty injection, the policy learned before novelty injection will likely to fail. On the other hand, if novelty injection makes a previously locked door unlocked, an agent that does not explore after novelty injection may always still seek out a key for a door that does not need it.
    \item \textbf{DoorKeyChange}: This novelty changes which key that opens a locked door. In MiniGrid doors are always unlocked by keys of the same color as the door. This means that if key and door colors do not match after novelty, agents will have to find another key to open the door. This may cause a previously learned policy to fail until the agent learns to start using the other key. 
    This novelty is illustrated in Figure~\ref{fig:splash}.
    \item \textbf{DoorNumKeys}: This novelty changes the number of keys needed to unlock a door. The default number of keys is one; this novelty tends to make policies fail because of the extra step of getting a second key.
    \item \textbf{ImperviousToLava}: Lava becomes non-harmful, whereas in Minigrid lava always immediately ends the episode with no reward. This may result in new routes to the goal that potentially bypass doors.
    \item \textbf{ActionRepetition}: This novelty changes the number of sequential timesteps an action will have to be repeated for it to occur. In MiniGrid it is usually assumed that for an action to occur it only needs to be issued once. So if an agent needed to command the pick-up action twice before novelty but only once afterwards, to reach its most efficient policy it would need to learn to not command pickup twice.
    \item \textbf{ForwardMovementSpeed} This novelty modifies the number of steps an agent takes each time the forward command is issued. In MiniGrid agents only move one gridsquare per time step. As a result, if the agent gets faster after novelty, the original policy may have a harder time controlling the agent, and will need to learn how to embrace this change that could make it reach the goal in fewer steps.
    \item \textbf{ActionRadius}: This novelty is an example of a change to the relational preconditions of an action by changing the radius around the agent where an action works. In MiniGrid this is usually assumed to be only a distance of one or zero, depending on the object. If an agent can pick up objects after novelty without being right next to them, it will have to realize this if it is to reach the optimum solution.
    \item \textbf{ColorRestriction}: This novelty restricts the objects one can interact with by color. In MiniGrid it is usually assumed that all objects can be interacted with. If an agent is trained with no blue interactions before novelty and then isn't allowed to interact with yellow objects after novelty, the agent will have to learn to pay attention to the color of objects.
    \item \textbf{Burdening}: This novelty changes the effect of actions based on whether the agent has any items in the inventory. In MiniGrid it is usually assumed that the inventory has no effect on actions. An agent experiencing this novelty, for example, might move twice as fast as usual when their inventory is empty, but half as fast as usual when in possession of the item, which it will have to compensate for strategically.
    \item \textbf{TransitionDeterminism}: This novelty changes the likelihood with which that actions selected by the agent occur. In MiniGrid it is usually assumed that all actions are deterministic. If an agent is trained with deterministic transitions before novelty and then experiences stochastic transitions after novelty, it will need to learn to take safe routes to the goal or its policy will fail more often
\end{itemize}

In Table~\ref{tbl:exemplars} we map each of the exemplar novelties to dimensions in our novelty ontology. 
To implement these novelties we also had to design custom versions of different standard MiniGrid objects, and these custom objects are also all included with \sysname.


\section{Evaluation and Baseline}

In novelty adaptation the core considerations by which we measure whether an agent adapted successfully involve not only performance on the task, but also the way the agent reacts to the novelty and the speed with which it recovers. To that end, we built the following metrics into \sysname:

\begin{itemize}
    \item {\em Resilience}: the difference in performance between a random agent and a pre-novelty agent when evaluated on the post-novelty domain {\em without} any adaptation.
    This represents the drop-off in performance when novelty is injected, relative to the performance of a random agent.
    A resilient agent may not encounter significant decrease in performance as in the case where the novelty is a goal location change and the agent has been trained on randomized grids.
    Barrier novelties may result in performance dropping to theoretical minimums.
    Shortcut novelties may result in no performance drop-off, but a random agent may experience greater reward.
    \item {\em Asymptotic adaptive performance}: final converged performance post novelty above random.
    This is what would simply be considered the converged performance of the agent in an environment with no novelty.
    \item {\em Adaptive efficiency}: the number of environment interactions required for post-novelty convergence.
    \item {\em One-shot adaptive performance}: the performance of the agent post-novelty after only one episode of interaction with the environment. 
\end{itemize}


\begin{figure}[t]
    \centering
    \includegraphics[width=\linewidth]{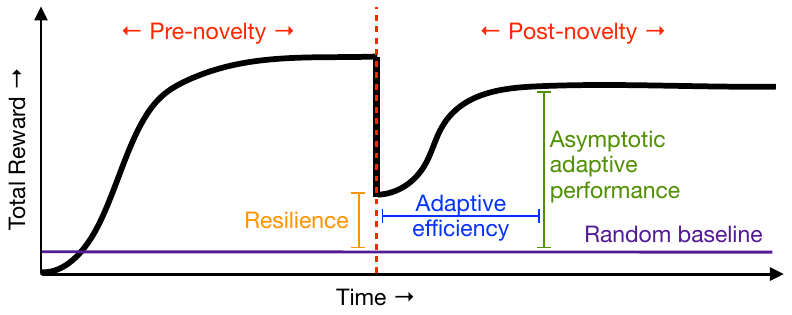}
    \caption{Several of the evaluation metrics illustrated against a notional performance curve for an agent.}
    \label{fig:eval}
\end{figure}

To demonstrate the way in which \sysname{} can be applied to the activity of sequential decision makers, we trained a reinforcement learning agent and the presence of novelty and measured its performance based on the metrics listed above. Specifically, we used the \texttt{DoorKeyToggle} novelty in an environment with 2 keys and 1 door on a 6x6 grid. It is set up so that before novelty injection the door is opened with one key and after it is opened with the other. 
The agent was trained as a proximal policy optimization deep reinforcement learning agent with a convolutional neural net feature extractor and two fully connected output networks, one to estimate the value and one to serve as the policy functions of the agent.

\begin{figure}[t]
    \centering
    \includegraphics[width=\linewidth]{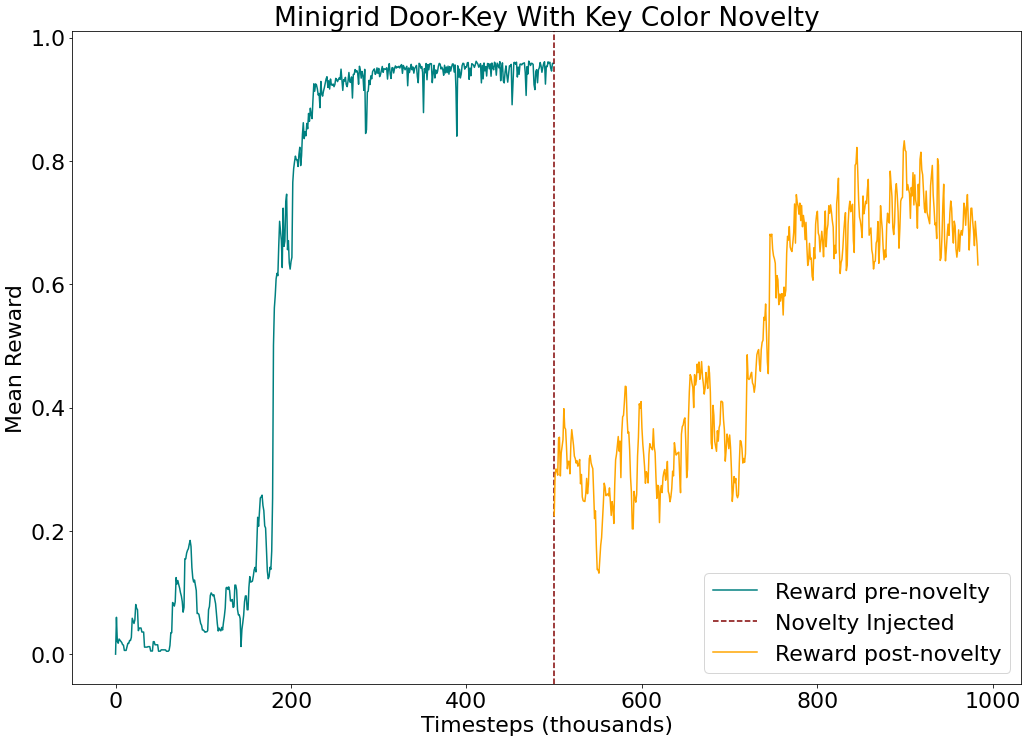}
    \caption{PPO baseline in \sysname{} using the \texttt{DoorKeyChange} novelty.
    The plot shows the progression of the agent learning, experiencing the novelty, and then adapting to the novelty.  
    The blue line indicates the learning process before novelty. 
    The novelty was injected at the 500k timestep, as indicated by the vertical red dotted line. 
    The yellow line shows adaptation to the novelty, which only converges around 300k timesteps after the novelty injection.
    This is expected as this baseline has no means of adaptation besides simply continuing to learn.
    }
    \label{fig:baselines}
\end{figure}

The agent was allowed to train for 500k time steps over just shy of 2000 episodes, at which point novelty was injected. As we can see in Figure \ref{fig:baselines} there is a precipitous drop in performance demarcating the novelty injection at this point. The agent is then able to train further for 500k time steps, yielding a more complete picture of the novelty adaptation of this agent.
This baseline agent has no novelty adaptability other than to continue learning using the extrinsic rewards that is provide by the environment. As a result, this represents a lower-bound against which to compare future novelty-adaptive agents. As illustrated in Figure~\ref{fig:baselines}, progress manifests itself as faster restoration of asymptotic maximum.

The results shown in Figure~\ref{fig:baselines} plot shows the progression of the agent learning from scratch in the original environment, experiencing the novelty, and then adapting to the novelty of the changed key. 
The random agent to which this agent is compared only every receives zero reward for this environment and task.
Examining the drop in performance where the novelty was injected at the 500k timestep (indicated by the vertical red dotted line) we can see low resilience of the baseline agent, with a resilience value of only 0.0531. 
Not visible on this map is the one-shot performance, or performance after one update over one full episode, which is actually reasonable at 0.22 . This tells us that the baseline reinforcement learning agent, while not efficient at getting to optimal post-novelty performance, has some promise as a starting point. 
Looking at the yellow line we can see the adaptive efficiency which only converges around 300k timesteps after the novelty injection, and the adaptive performance converges to a lower 0.8 reward.
This from this we can tell that the agent is not effective at adapting, but expected as this baseline has no means of adaptation besides simply continuing to learn.

Given that PPO is an algorithm that is near-state of the art in many reinforcement learning tasks, this experiment serves as an important demonstration of how much more needs to be researched in novelty adaptation. Future solutions to novelty adaptation can seek better resilience, adaptive efficiency, asymptotic adaptive performance, or even one-shot adaptive performance as there is much room for improvement in all of these areas. 


\section{Future Work}

Novelty in sequential decision making is a rich space that promises to enhance robustness of agents in virtual worlds in anticipation of operation in the real world. 
There are a number of ways that the ontology and the way it manifests itself in \sysname{} can be enhanced in future iterations as research in this space matures.
When it comes to novelties, the two major axes of novelty undressed by this work have to do with the (a)~local or global application of novelties and (b)~populations of agents, including the behaviors of external agents. When we differentiate local and global novelties, we mean to say that novelty can affect individual entities or instances as well as any entities or instances of a certain type or class; agents will react differently if a novelty changes the way all doors operate as opposed to the way one door operates. 
However, we have not yet factored that dimension into \sysname.

Observation of other agents performing the same or related task can have implications on novelty adaptation.
When agents are acting in the presence of other agents it has a powerful effect on the long-term performance of the agent as well as the learning ability of the agent. For example, if agents are competing for the same resources to reach the same goal this affects the strategy agents will take to reach that goal, and agents can use other agents effectively as a source of exploration when external agents do something that it originally thought was not possible.
Indeed, this may be a significant way in which novelty-adaptive agents detect and adapt to shortcut novelties.

Another way that other agents factor into novelty is in adversarial settings where the novelty may be a change in the behavior or strategy of the adversary, up to and including adversaries becoming cooperative or vice versa.

Beyond and expanded novelty ontology, additional measurement and quantification of novelty is an important future direction for \sysname. Measuring the difference between these distributions of novelties is key among these measures as allows comparison between different novelties. There are many ways to quantify differences in distribution, common among them Shannon Jensen divergences like KL divergence, and metrics like the earth-movers distance. Additionally, very recent work has examined using fixed agent baselines to characterize differences in distribution, as well as metrics of mutual information and edit distance. Integrating metrics like these would be extremely valuable additions to \sysname{} as it would enable researchers to not only compare these novelties, but also to set expectations of novelty adaptation based on the distribution differences. Along this same thread integrating metrics of novelty detection and characterization into \sysname{} may be of great interest to want to study these subproblems in the context of sequential decision making problems. 

\section{Acknowledgements}

This research is sponsored in part by the Defense Advanced Research Projects Agency (DARPA), under contract number W911NF-20-2-0008. Views and conclusions contained in this document are those of the authors and should not be interpreted as necessarily representing official policies or endorsements, either expressed or implied, of the US Department of Defense or the United States Government.
We gratefully acknowledge David Aha, Katarina Doctor, and members of the DARPA SAIL-ON Working Group on Metrics.

\bibliography{references}
\end{document}